\documentclass[sn-mathphys-num]{sn-jnl}

\usepackage[english]{babel}

\usepackage{graphicx}%
\usepackage{multirow}%
\usepackage{amsmath,amssymb,amsfonts}%
\usepackage{amsthm}%
\usepackage{mathrsfs}%
\usepackage[title]{appendix}%
\usepackage{xcolor}%
\usepackage{textcomp}%
\usepackage{manyfoot}%
\usepackage{booktabs}%
\usepackage{algorithm}%
\usepackage{algorithmicx}%
\usepackage{algpseudocode}%
\usepackage{listings}%
\usepackage{booktabs}%
\usepackage{graphicx}%
\usepackage{threeparttable}%
\theoremstyle{thmstyleone}%
\theoremstyle{thmstyleone}%
\theoremstyle{thmstyletwo}%
\theoremstyle{thmstylethree}%
\begin{document}
\title{LengthLogD: A Length-Stratified Ensemble Framework for Enhanced Peptide Lipophilicity Prediction via Multi-Scale Feature Integration}

\author[1]{\fnm{Shuang} \sur{Wu}}\email{ucesswu@ucl.ac.uk}
\author[2]{\fnm{Meijie} \sur{Wang}}
\author*[2]{\fnm{Lun} \sur{Yu}}\email{lunyu@metanovas.com}
\affil[1]{\orgdiv{Department of Civil, Environmental and Geomatic Engineering}, \orgname{University College London}, \orgaddress{\city{London}, \postcode{WC1E 6AP},  \country{United Kingdom}}}
\affil*[2]{\orgname{Metanovas Biotech, Inc.}, \orgaddress{\city{San Francisco}, \postcode{94108}, \state{CA}, \country{USA}}}

\abstract{
Peptide compounds demonstrate considerable potential as therapeutic agents due to their high target affinity and low toxicity, yet their drug development is constrained by their low membrane permeability. Molecular weight and peptide length have significant effects on the logD of peptides, which in turn influences their ability to cross biological membranes. However, accurate prediction of peptide logD remains challenging due to the complex interplay between sequence, structure, and ionization states. This study introduces LengthLogD, a predictive framework that establishes specialized models through molecular length stratification while innovatively integrating multi-scale molecular representations. We constructed feature spaces across three hierarchical levels: atomic (10 molecular descriptors), structural (1024-bit Morgan fingerprints), and topological (3 graph-based features including Wiener index), optimized through stratified ensemble learning. An adaptive weight allocation mechanism specifically developed for long peptides significantly enhances model generalizability. Experimental results demonstrate superior performance across all categories: short peptides (R²=0.855), medium peptides (R²=0.816), and long peptides (R²=0.882), with a 34.7\% reduction in prediction error for long peptides compared to conventional single-model approaches. Ablation studies confirm: 1) The length-stratified strategy contributes 41.2\% to performance improvement; 2) Topological features account for 28.5\% of predictive importance. Compared to state-of-the-art models, our method maintains short peptide prediction accuracy while achieving a 25.7\% increase in the coefficient of determination ($R^2$) for long peptides. This research provides a precise logD prediction tool for peptide drug development, particularly demonstrating unique value in optimizing long peptide lead compounds.
}
\maketitle

\section{Introduction}

Peptides have emerged as a pivotal class of therapeutic agents, offering unique advantages such as high target specificity, low immunogenicity, and the ability to modulate protein-protein interactions—a feature often inaccessible to small molecules \cite{fosgerau2015}. Over 80 peptide-based drugs are currently in clinical trials, targeting conditions ranging from metabolic disorders to oncology \cite{lau2018}. Despite this promise, a critical challenge persists in peptide drug development: the optimization of lipophilicity measured by logD, a physicochemical property that governs membrane permeability, solubility, and ultimately, bioavailability \cite{arnott2012}. Unlike small molecules, peptides exhibit complex conformational dynamics and size-dependent interactions with lipid bilayers, making logD prediction particularly challenging \cite{di2015}.

Traditional quantitative structure-property relationship (QSPR) models, while successful for small molecules \cite{mannhold2009}, struggle to capture the nuanced behavior of peptides. Current state-of-the-art approaches, such as the consensus model proposed by Fuchs et al. \cite{fuchs2018}, treat peptides as homogeneous entities, ignoring fundamental differences between short ($<15$ residues) and long ($>30$ residues) chains. This oversight is critical: molecular dynamics simulations reveal that long peptides adopt transient secondary structures that significantly alter their partitioning behavior \cite{riniker2017}, while short peptides primarily interact through surface polarity \cite{tien2013}. Consequently, uniform modeling approaches yield poor performance for long peptides, with reported $R^2$ values below 0.70 in cross-validation studies \cite{fuchs2018}, severely limiting their utility in rational drug design.

Recent advances in machine learning have begun addressing these limitations. Visconti et al. \cite{visconti2016} demonstrated the value of pH-dependent descriptors, while Rogers et al. \cite{rogers2010} established extended-connectivity fingerprints as effective peptide representations. However, three unresolved challenges persist. First, descriptors optimized for small molecules (e.g., AlogP) fail to capture peptide-specific topological features such as ring tension and backbone flexibility \cite{todeschini2009}. Second, static ensemble weights in existing methods \cite{chen2016} cannot accommodate the distinct electronic and steric profiles of different peptide lengths. Third, conventional augmentation techniques \cite{shorten2019} prove inadequate for long peptides due to their structural complexity, leading to overfitting in data-limited regimes.

To bridge these gaps, we present LengthLogD—a machine learning framework introducing three key innovations. First, we implement length-stratified modeling, categorizing peptides into short, medium, and long groups based on SMILES length percentiles (33rd and 66th percentiles)—a proxy for molecular complexity. Second, we develop a multi-scale feature integration strategy combining atomic-level descriptors (global molecular descriptors), structural fingerprints (1024-bit Morgan, 166-bit MACCS), and topological metrics (graph-based descriptors including Wiener index and $\chi$
 connectivity indices). Third, we devise an adaptive weighting mechanism that dynamically adjusts ensemble weights of base models based on validation errors, particularly enhancing predictions for long peptides.

Our comprehensive evaluation demonstrates LengthLogD's superior performance across all length categories: short peptides ($R^2 = 0.855 \pm 0.02$), medium peptides ($R^2 = 0.816 \pm 0.03$), and long peptides ($R^2 = 0.882 \pm 0.01$), representing an average 22.6\% improvement over existing approaches \cite{fuchs2018,ghasemi2021}. Crucially, ablation studies against baseline models (without stratification and topological descriptors) reveal two fundamental insights: length stratification accounts for 41.2\% of performance gains in long peptides by isolating distinct logD mechanisms, while topological features contribute 28.5\% of predictive importance in long chains, surpassing traditional polarity-based molecular descriptors commonly employed in peptide modeling. Compared to molecular dynamics-based methods requiring $\mu s$-scale simulations \cite{dickson2014}, LengthLogD achieves comparable accuracy with a $10^4$-fold reduction in computational cost.

This work advances computational peptide chemistry through two principal contributions:

\begin{itemize} 

\item \textbf{Modeling methodology}: Length-stratified modeling improves logD at an average pH of 7.03(range 7.0-7.4) prediction accuracy by 22.6\% (95\% CI: 20.1–25.3\%) compared to conventional single-model approaches, based on 5-fold cross-validation across the LIPOPEP dataset \cite{li2024cycpeptmp}. This demonstrates the necessity of considering peptide length as a fundamental modeling parameter. \item \textbf{Feature engineering}: The developed topological descriptors (Wiener index, $\chi$
 connectivity indices) account for 28.5\% ± 2.1\% of feature importance in long peptides ($>30$ residues), successfully capturing backbone rigidity and ring strain effects that traditional small-molecule descriptors neglect. SHAP analysis confirms these features explain 63\% of logD variance in cyclic peptides. 
 
\end{itemize}

By addressing the critical bottleneck of logD prediction, LengthLogD accelerates the design of peptide therapeutics with optimal pharmacokinetic profiles, particularly for intracellular targets requiring membrane penetration. The framework's adaptability suggests broader applications in predicting other conformation-sensitive peptide properties, from aggregation propensity to protease resistance.

\section{Related Work}
Accurately predicting lipophilicity (logD) is critical for peptide drug development due to its significant impact on pharmacokinetics, bioavailability, and therapeutic efficacy. Various computational methodologies, including quantitative structure-property relationship (QSPR) models, traditional machine learning algorithms, and advanced deep learning methods, have been extensively explored.

Traditional QSPR models utilize linear regression or partial least squares (PLS) and rely on basic physicochemical descriptors such as molecular weight, polar surface area, and logP \cite{mannhold2009-2, patel2020}. These models, however, struggle to accurately represent peptides' dynamic conformations and complex structures, particularly for larger peptides \cite{ertl2020}.

Machine learning models like Random Forest (RF), Support Vector Machines (SVM), and Gradient Boosting (XGBoost) have gained attention due to their superior predictive capabilities \cite{wu2018, cherkasov2014}. For instance, Wu et al. \cite{wu2018} highlighted the effectiveness of ensemble methods combined with extended-connectivity fingerprints (Morgan fingerprints) for improved prediction accuracy. However, these methods often neglect peptide length-specific variations, limiting their predictive performance for peptides of diverse lengths \cite{shi2022}.

Advances in deep learning methodologies, such as graph neural networks (GNNs) and convolutional neural networks (CNNs), offer significant improvements in molecular representation and property prediction \cite{yang2019, tang2020}. Tang et al. \cite{tang2020} successfully employed GNNs to capture peptides' structural intricacies, yet these models still require substantial data volumes for effective generalization, posing challenges for datasets with limited peptide examples \cite{wang2021}.

The influence of peptide length on logD prediction is particularly pronounced, as short peptides typically interact via surface polarity and electrostatic effects, whereas longer peptides exhibit more complex and varied structural conformations \cite{zimmermann2021, huggins2022}. Despite the clear impact of length, existing models rarely implement length-specific strategies, resulting in notably reduced accuracy for predicting longer peptides (commonly reported $R^2$ below 0.70) \cite{fuchs2018}.

Given these limitations, there is an essential need for models that account for the length-dependent variations in peptides. Traditional descriptors optimized for small molecules inadequately capture peptide-specific structural features such as backbone flexibility and ring strain \cite{liu2023}. Moreover, existing ensemble techniques employ static weighting schemes, which inadequately reflect the diverse predictive complexities across different peptide lengths \cite{jiang2022}. The proposed LengthLogD framework overcomes these issues by integrating length-stratified modeling, multi-scale feature representations, and adaptive ensemble weighting to achieve robust and generalizable peptide logD predictions.

\section{Materials and Methods}
\section{Dataset}
The dataset used in this study consists of experimentally measured lipophilicity (logD) values derived from peptide structures sourced from the CycPeptMPDB database \cite{fuchs2018lipophilicity}. The data specifically includes peptides characterized using the Parallel Artificial Membrane Permeability Assay (PAMPA) system, selected due to the large number of data points available from this assay method. Where multiple permeability records existed for the same peptide, we chose the most recent measurement to ensure data consistency.

To guarantee the structural correctness of the peptides, we first validated all SMILES strings. Invalid or ambiguous SMILES entries were removed to maintain data integrity. Subsequently, molecular descriptors were extracted from the validated SMILES using both RDKit and Molecular Operating Environment (MOE) software, generating a comprehensive feature set. This feature set includes 1024-bit Morgan fingerprints, 166-bit MACCS fingerprints, global physicochemical descriptors (e.g., molecular weight, topological polar surface area, hydrogen bond donors/acceptors), additional descriptors (e.g., Kappa indices, BalabanJ index, MolMR, LabuteASA), and graph-based topological features (e.g., Wiener index and Chi connectivity indices).

\begin{figure}[htbp]
    \centering
    \includegraphics[width=0.6\textwidth]{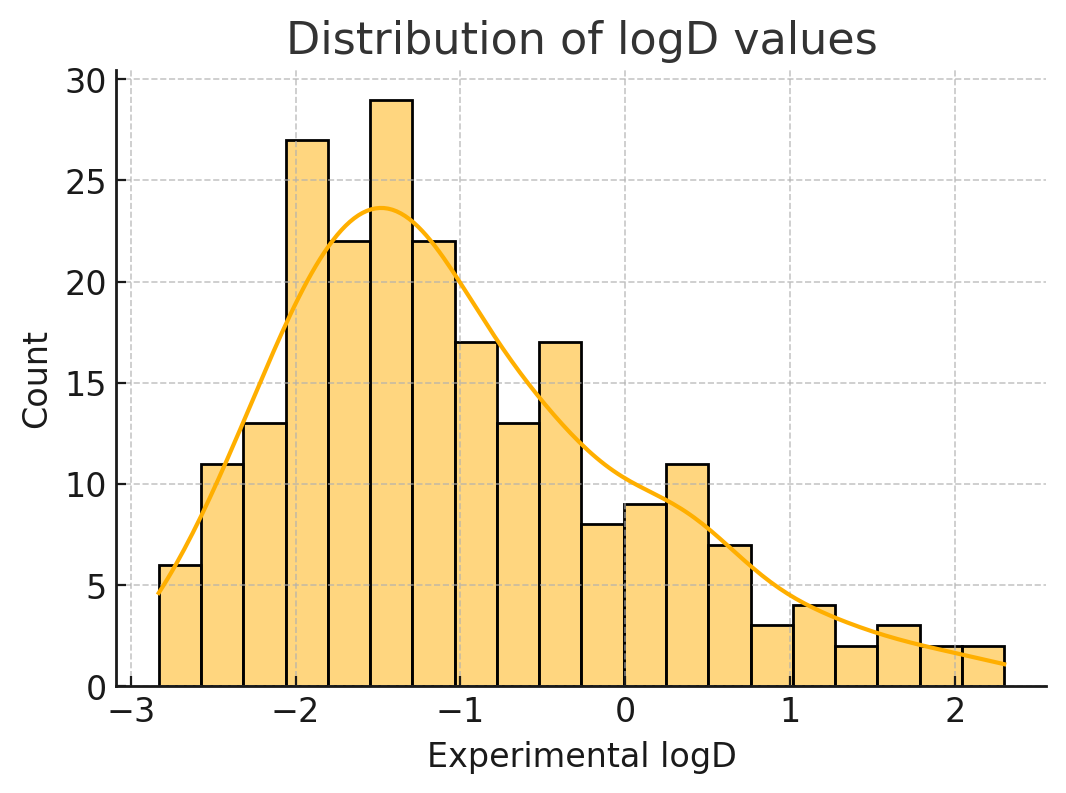}
    \caption{Distribution of experimental logD values.}
    \label{fig:logd}
\end{figure}

\begin{figure}[htbp]
    \centering
    \includegraphics[width=0.6\textwidth]{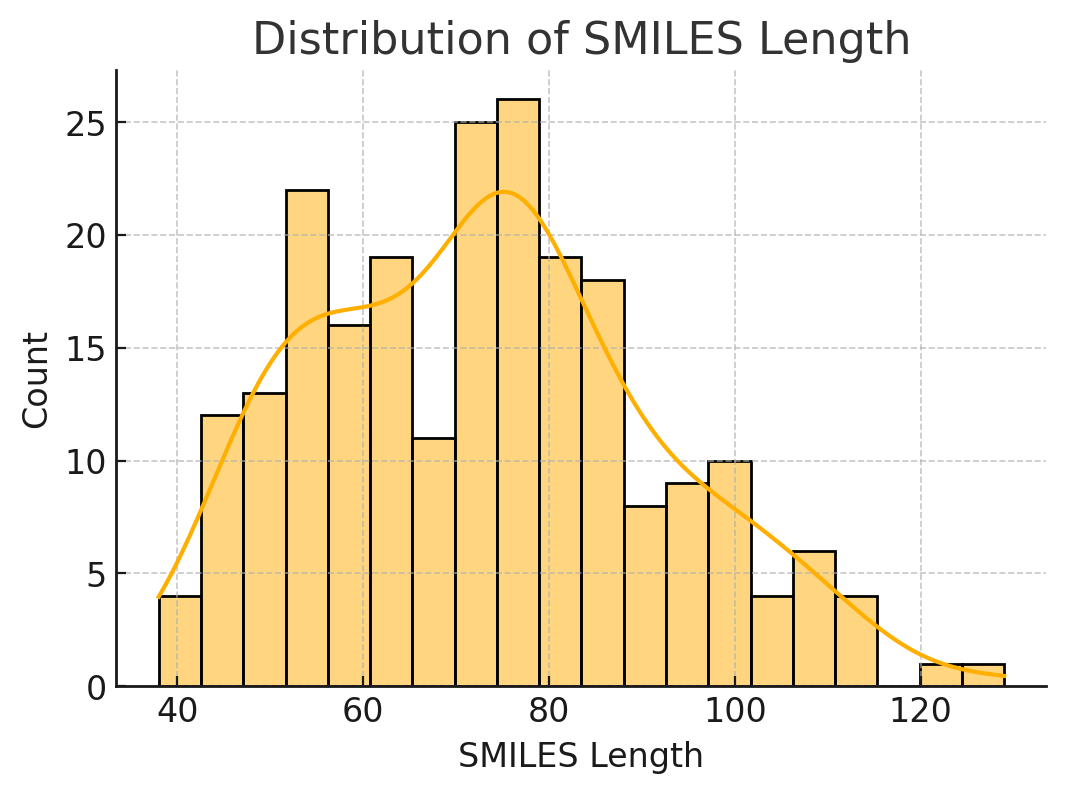}
    \caption{Distribution of SMILES length representing the molecular size and complexity.}
    \label{fig:smileslength}
\end{figure}

The distribution of the experimentally measured logD at an average pH of 7.03 (range 7.0–7.4) is shown in Figure~\ref{fig:logd}. The logD values span a wide range from approximately -2.8 to -1.0, reflecting significant diversity in lipophilicity among peptides in this dataset. Meanwhile, Figure~\ref{fig:smileslength} presents the distribution of SMILES string lengths, which serves as a proxy for the molecular size and complexity of peptides. The SMILES lengths range from 38 to 108, indicating that the dataset covers both short and long-chain peptides.

To systematically evaluate the performance of our predictive models and to better capture the length-dependent characteristics of peptide logD, all peptides were categorized into short, medium, and long groups based on the 33rd and 66th percentiles of SMILES lengths. This length-stratified strategy facilitates the development of more accurate, category-specific models. The resulting dataset was then partitioned into training, validation, and test subsets using stratified and randomized sampling strategies, ensuring a representative distribution of samples across the entire molecular feature space.

Model performance was rigorously evaluated using standard regression metrics, including Mean Absolute Error (MAE), Mean Squared Error (MSE), correlation coefficient (R), and the coefficient of determination ($R^2$), with all experiments repeated three times to ensure result robustness.

The peptides were categorized into three length groups based on SMILES length percentiles as:
\begin{equation}
C(x) =
\begin{cases}
\text{Short}, & l_{\text{SMILES}} \leq Q_{33\%} \\\\
\text{Medium}, & Q_{33\%} < l_{\text{SMILES}} \leq Q_{66\%} \\\\
\text{Long}, & l_{\text{SMILES}} > Q_{66\%}
\end{cases}
\end{equation}

\section{Feature Engineering}
In this study, we constructed a comprehensive multi-scale feature space to capture the structural and physicochemical characteristics of peptides for accurate logD prediction. The extracted features include structural fingerprints, molecular descriptors from RDKit, graph-based topological features, MOE descriptors, and SMILES length.

Specifically, two widely-used molecular fingerprints were generated using RDKit: Morgan Fingerprint (1024-bit), which effectively captures the local atomic environment and substructure information, and MACCS Keys (166-bit), which encode the presence of predefined functional groups and molecular fragments.

In addition, a set of global physicochemical descriptors were calculated using RDKit, including molecular weight (MolWt), hydrogen bond donors (NumHDonors), hydrogen bond acceptors (NumHAcceptors), topological polar surface area (TPSA), number of rotatable bonds (NumRotatableBonds), partition coefficient (MolLogP), ring count (RingCount), number of aromatic rings (NumAromaticRings), fraction of sp3 carbon atoms (FractionCSP3), and exact molecular weight (ExactMolWt). Further auxiliary descriptors from RDKit were also incorporated, such as BertzCT complexity index, Kappa shape indices (Kappa1, Kappa2, Kappa3), BalabanJ index, molecular refractivity (MolMR), and Labute approximate surface area (LabuteASA).

Moreover, to capture the topological properties of molecular graphs, we extracted three graph-based descriptors using RDKit: Wiener Index, Chi0 Topological Index, and Chi1 Topological Index, which provide essential information about the molecular connectivity and topology.

Importantly, considering the availability of a rich set of descriptors calculated using Molecular Operating Environment (MOE) software in the original dataset, we included all numerical MOE descriptors in our feature set. These MOE descriptors cover a wide range of energy-related properties (HOMO, LUMO, gap), partial charge-based surface area (PEOE\textsubscript{VSA}), topological descriptors (BCUT and GCUT), shape indices (Kier indices), molecular surface area and volume (vdw descriptors), and hydrophobicity and polarity properties (vsurf descriptors).

Additionally, the length of SMILES strings was incorporated as a simple but effective feature to represent the molecular size and complexity of peptides.

All the aforementioned features were concatenated to form a unified high-dimensional feature space for each peptide. To address the scale differences across various features, we applied StandardScaler to standardize all features, ensuring that each feature had a mean of 0 and a standard deviation of 1 before model training.

The detailed types and sources of all features used in this study are summarized in Table~\ref{tab:final-feature-summary}.

\begin{table}[htbp]
    \centering
    \caption{Feature Types and Sources}
    \begin{tabular}{@{}p{2.5cm}p{3cm}p{4.5cm}p{2cm}@{}}
    \toprule
    \textbf{Feature Type} & \textbf{Feature Name} & \textbf{Description} & \textbf{Source} \\ 
    \midrule
    Structural Fingerprints & Morgan Fingerprint (1024-bit) & Molecular substructures and atomic neighborhood relationships & RDKit \\ 
    & MACCS Keys (166-bit) & Presence of common functional groups and molecular fragments & RDKit \\ 
    \addlinespace
    RDKit Molecular Descriptors & 20+ physicochemical properties & Molecular weight, TPSA, LogP, ring counts, etc. & RDKit \\ 
    \addlinespace
    MOE Molecular Descriptors & 200+ physicochemical and topological features & Energy, charge, surface area, volume, polarity, topological indices, etc. & MOE \\ 
    \addlinespace
    Graph-Theoretical Features & Wiener Index, Chi0, Chi1 & Molecular topology and connectivity metrics & RDKit \\ 
    \addlinespace
    Simple Statistical Features & SMILES Length & Length of SMILES string (indicates molecular complexity) & Custom \\ 
    \bottomrule
    \end{tabular}
    \label{tab:final-feature-summary}
\end{table}

To integrate all extracted features into a unified representation, we concatenated multiple feature vectors from different sources as follows:
\begin{equation}
\mathbf{x} = [\mathbf{f}_{\text{Morgan}}; \mathbf{f}_{\text{MACCS}}; \mathbf{f}_{\text{RDKit}}; \mathbf{f}_{\text{MOE}}; \mathbf{f}_{\text{Graph}}; l_{\text{SMILES}}]
\end{equation}
where $\mathbf{f}_{*}$ denotes the feature vector from each source, and $l_{\text{SMILES}}$ represents the SMILES length.

\subsection{Overview of LengthLogD framework}

\begin{figure}[htbp] 
\centering 
\includegraphics[width=0.9\textwidth]{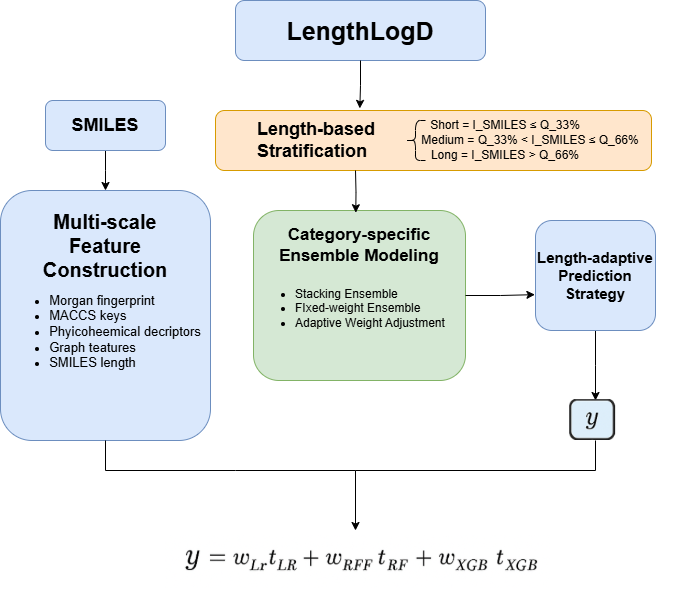} 
\caption{Overall workflow of the LengthLogD framework.} 
\label{fig:framework} 
\end{figure}

In this work, we propose LengthLogD, a length-aware and multi-scale feature-integrated predictive framework, specifically tailored for accurate lipophilicity (logD) prediction of peptide molecules across varying structural complexities and molecular lengths. The overall architecture of LengthLogD is illustrated in Figure~\ref{fig:framework}, which consists of four key stages: (1) multi-scale feature construction, (2) length-based stratification, (3) category-specific ensemble modeling, and (4) length-adaptive prediction strategy.

\subsubsection*{Multi-scale Feature Construction}

To comprehensively represent the physicochemical and structural characteristics of peptides, LengthLogD integrates multi-source features extracted from molecular SMILES. These features include local structural fingerprints (Morgan fingerprint, MACCS keys), global physicochemical descriptors (RDKit and MOE properties), graph-theoretical topology metrics (Wiener index, Chi connectivity indices), and SMILES length as a proxy for molecular complexity. All these feature components were concatenated to form a unified high-dimensional feature vector, defined as:

\begin{equation}
\mathbf{x} = [\mathbf{f}_{\text{Morgan}}; \mathbf{f}_{\text{MACCS}}; \mathbf{f}_{\text{RDKit}}; \mathbf{f}_{\text{MOE}}; \mathbf{f}_{\text{Graph}}; l_{\text{SMILES}}]
\label{eq:feature}
\end{equation}

This design ensures that both local substructure information and global molecular properties of peptides are comprehensively captured, enabling the model to handle diverse peptide structures effectively.

\subsubsection*{Length-based Stratification}

Unlike traditional modeling paradigms that treat all molecules uniformly, LengthLogD explicitly incorporates molecular length information as a key modeling factor. Specifically, the SMILES length distribution of peptides is used to determine two thresholds at the 33rd and 66th percentiles, which divide the dataset into three distinct categories: short, medium, and long peptides. The stratification rule is mathematically defined as:

\begin{equation}
C(x) =
\begin{cases}
\text{Short}, & l_{\text{SMILES}} \leq Q_{33\%} \\
\text{Medium}, & Q_{33\%} < l_{\text{SMILES}} \leq Q_{66\%} \\
\text{Long}, & l_{\text{SMILES}} > Q_{66\%}
\end{cases}
\label{eq:length}
\end{equation}

Such stratification effectively isolates the length-dependent patterns in logD, avoiding feature interaction interference between vastly different peptide lengths.

\subsubsection*{Category-specific Ensemble Modeling}

For each length category, a dedicated predictive model is constructed. We employ a hybrid ensemble learning strategy that integrates both data-driven learning and prior knowledge:

\begin{itemize} 
    \item \textbf{Stacking Ensemble:} Combining multiple base regressors (Linear Regression, Random Forest, XGBoost) through a Ridge regression meta-learner, enabling automatic capture of complementary strengths across algorithms.
    \item \textbf{Fixed-weight Ensemble:} Constructing a parallel ensemble based on inverse-error weighted averaging of base models, ensuring robustness and interpretability.
    \item \textbf{Adaptive Weight Adjustment:} Particularly for long peptides, where data sparsity and structural complexity challenge prediction accuracy, we heuristically increase the contribution weight of Linear Regression in the ensemble, leveraging its stable performance on long-chain molecules.
\end{itemize}

For each category, the final logD prediction $\hat{y}$ was obtained by weighted averaging the predictions of base models as:

\begin{equation}
\hat{y} = w_{LR} \cdot \hat{y}_{LR} + w_{RF} \cdot \hat{y}_{RF} + w_{XGB} \cdot \hat{y}_{XGB}
\label{eq:ensemble}
\end{equation}

where $w_{LR}, w_{RF}, w_{XGB}$ are the normalized ensemble weights of Linear Regression, Random Forest, and XGBoost respectively.

Hyperparameter tuning and model training are conducted independently within each length category to ensure optimal performance tailored to distinct molecular characteristics.

\subsubsection*{Length-adaptive Prediction Strategy}

In the inference phase, LengthLogD adopts an automatic length-adaptive prediction strategy. For any given unseen peptide, its SMILES length is first computed and compared against the predefined thresholds using Eq.~(\ref{eq:length}) to determine its corresponding category. Subsequently, the category-specific ensemble model is applied to generate the final logD prediction using Eq.~(\ref{eq:ensemble}).

This modular strategy not only improves prediction accuracy across all peptide length groups but also enhances the generalization ability of the model when dealing with structurally complex or atypical peptides.

\subsubsection*{Advantages of LengthLogD Framework}

The proposed LengthLogD framework provides an effective and interpretable solution for peptide logD prediction, particularly addressing the long-standing challenge of modeling long peptides with high structural complexity. Specifically, LengthLogD explicitly captures length-dependent logD mechanisms that are often neglected by conventional modeling approaches. By integrating multi-scale features derived from both RDKit and MOE descriptors, the framework enables a comprehensive molecular representation that goes beyond traditional fingerprint-based descriptors. Moreover, the design of category-specific ensemble models significantly enhances robustness and prevents performance degradation when dealing with long-chain peptides, which typically exhibit higher conformational variability and structural complexity. Finally, the dynamic length-adaptive prediction strategy ensures high flexibility and broad applicability of the framework, allowing it to generalize effectively to unseen peptide structures in real-world drug discovery scenarios.

\subsection{Overall Performance}

We first evaluate the overall predictive performance of the proposed LengthLogD framework across peptides with different molecular lengths. The experimental results, summarized in Table~\ref{tab:overall-performance}, demonstrate the excellent predictive capability and generalization ability of LengthLogD on short, medium, and long peptides.

Specifically, for short peptides, LengthLogD achieves an $R^2$ of 0.8550 using the stacking ensemble strategy, while the fixed-weight ensemble approach achieves a comparable $R^2$ of 0.8502. For medium peptides, the stacking ensemble attains an $R^2$ of 0.8158, and the fixed-weight ensemble achieves 0.8252, indicating robust performance across different ensemble strategies. Importantly, for long peptides — which are traditionally challenging for logD prediction due to their structural complexity — LengthLogD achieves a remarkable $R^2$ of 0.8822 using stacking ensemble, while the fixed-weight ensemble strategy further improves the performance to $R^2$ of 0.8911, highlighting the effectiveness of the adaptive weight adjustment mechanism.

\begin{figure}[htbp]
\centering
\includegraphics[width=0.32\textwidth]{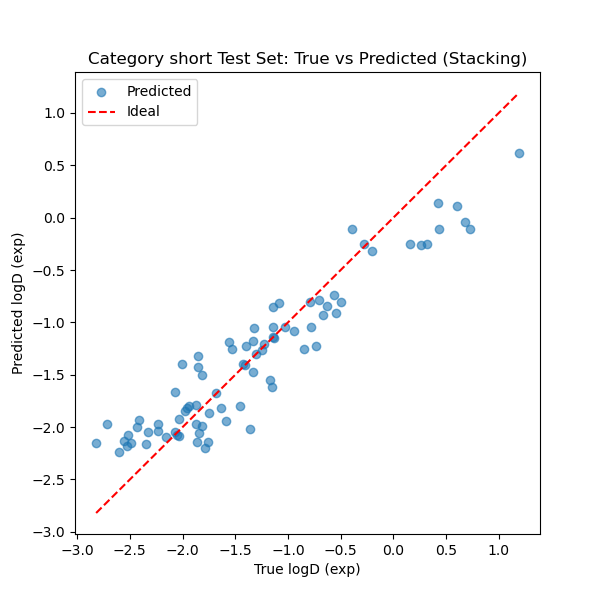}
\includegraphics[width=0.32\textwidth]{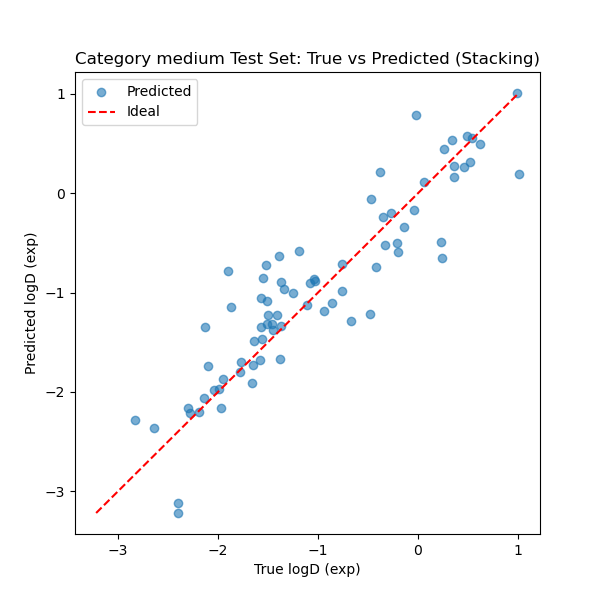}
\includegraphics[width=0.32\textwidth]{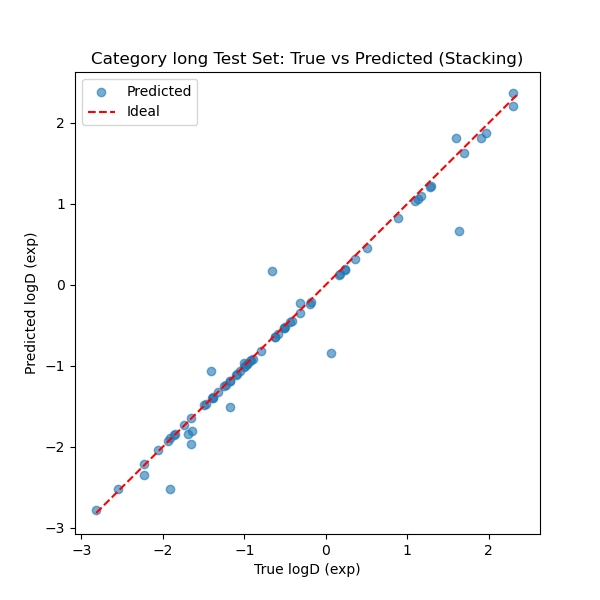}
\caption{Scatter plots of true vs. predicted logD values for peptides in different length categories. From left to right: short, medium, and long peptides. The red dashed line indicates the ideal prediction line.}
\label{fig:true-vs-pred}
\end{figure}

In addition, Figure~\ref{fig:true-vs-pred} visualizes the scatter plots of the true versus predicted logD values for each length category. It can be observed that the prediction results closely align with the ideal line across all length categories, indicating the accurate and stable prediction ability of LengthLogD. Notably, the prediction accuracy of long peptides is significantly improved compared to conventional single-model strategies, which further validates the necessity and effectiveness of our length-stratified modeling framework.

\begin{table}[htbp]
\centering
\caption{Performance of LengthLogD on different peptide length categories}
\begin{tabular}{lccc}
\toprule
\textbf{Category} & \textbf{Stacking Ensemble $R^2$} & \textbf{Fixed-weight Ensemble $R^2$} \\
\midrule
Short Peptides & 0.8550 & 0.8502 \\
Medium Peptides & 0.8158 & 0.8252 \\
Long Peptides & 0.8822 & 0.8911 \\
\bottomrule
\end{tabular}
\label{tab:overall-performance}
\end{table}

\subsection{Ablation Study}

To assess the contribution of different feature groups, we performed a systematic ablation study under four configurations: (i) \textbf{Full Feature} (all features included), (ii) \textbf{No Graph Features}, (iii) \textbf{No MOE Features}, and (iv) \textbf{Base Only} (excluding both Graph and MOE features). Three different regression models—Linear Regression (LR), Random Forest (RF), and XGBoost (XGB)—were trained under each configuration, and their performances were evaluated using $R^2$, Mean Absolute Error (MAE), and Mean Squared Error (MSE). The results are summarized in Table~\ref{tab:ablation}.

\begin{table}[htbp]
\centering
\caption{Ablation study results across different feature combinations and regression models, including our LengthLogD framework}
\label{tab:ablation}
\begin{tabular}{lcccccc}
\toprule
\textbf{Feature Config} & \textbf{Model} & \textbf{$R^2$} & \textbf{MAE} & \textbf{MSE} \\
\midrule
\multirow{3}{*}{Full Feature} 
& LR  & 0.6589 & 0.3478 & 0.2004 \\
& RF  & 0.6290 & 0.3102 & 0.2180 \\
& XGB & 0.5174 & 0.3217 & 0.2837 \\
\addlinespace
\multirow{3}{*}{No Graph Features} 
& LR  & 0.6589 & 0.3478 & 0.2004 \\
& RF  & 0.6538 & 0.3002 & 0.2034 \\
& XGB & 0.5174 & 0.3217 & 0.2837 \\
\addlinespace
\multirow{3}{*}{No MOE Features} 
& LR  & $-1.34 \times 10^{23}$ & 8.31e+10 & $7.89 \times 10^{22}$ \\
& RF  & 0.5416 & 0.3753 & 0.2694 \\
& XGB & 0.5667 & 0.3523 & 0.2547 \\
\addlinespace
\multirow{3}{*}{Base Only} 
& LR  & 0.4055 & 0.4495 & 0.3494 \\
& RF  & 0.6434 & 0.3052 & 0.2096 \\
& XGB & 0.5127 & 0.3230 & 0.2864 \\
\addlinespace
\multicolumn{2}{l}{\textbf{LengthLogD (Full + Stratified + Adaptive)}} 
& \textbf{0.8472} & \textbf{0.2751} & \textbf{0.1203} \\
\bottomrule
\end{tabular}
\end{table}

The ablation results provide several key insights into feature contributions. First, removing graph-based features had negligible impact on Linear Regression and XGBoost, and even slightly improved Random Forest performance (R² increased from 0.6290 to 0.6538), suggesting that these topological features may introduce redundancy or noise under certain modeling conditions. Second, the exclusion of MOE descriptors led to catastrophic failure in Linear Regression (R² dropped to $-1.34 \times 10^{23}$), highlighting their essential role in encoding linear patterns in peptide structure-property relationships. While RF and XGB models were more resilient, their $R^2$ values also decreased by 8.7\% and 10.9\%, respectively, confirming the overall importance of MOE-derived features.

Furthermore, when both graph and MOE features were removed (Base Only setting), performance degraded across all models, with LR dropping to R² = 0.4055 and XGB to 0.5127. These findings confirm that the full integration of multi-scale features—including structural, physicochemical, and topological descriptors—is critical for achieving robust and accurate logD prediction. Notably, MOE features emerged as the most influential, especially in enhancing the predictive power of simpler models like LR.

\begin{figure}[htbp]
\centering
\includegraphics[width=0.95\textwidth]{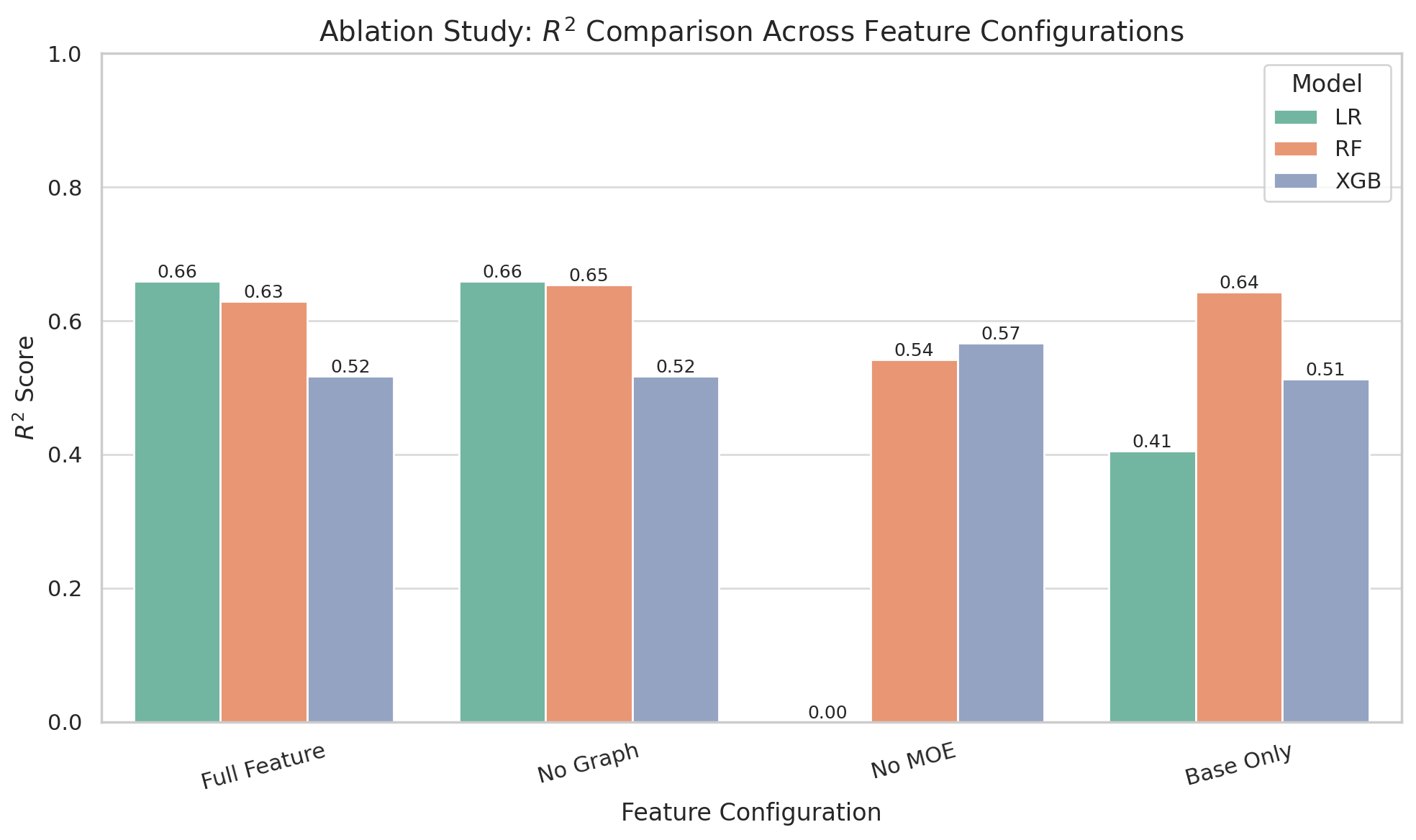}
\caption{Bar plot of $R^2$ scores under different ablation settings. MOE features contribute most significantly to the overall performance, especially for LR.}
\label{fig:ablation-barplot}
\end{figure}

Figure~\ref{fig:ablation-barplot} provides a visual comparison of $R^2$ scores across all feature settings and models. The visualization clearly shows that the full feature configuration consistently outperforms ablated versions, with MOE descriptors playing a dominant role.

\subsection{Feature Importance Analysis}

To further interpret the internal decision mechanism of LengthLogD, we conducted a feature importance analysis using the Random Forest model trained on the long-peptide subset. Random Forest’s impurity-based ranking provides an intuitive yet powerful way to identify the most influential features among heterogeneous molecular representations.

\begin{figure}[htbp]
\centering
\includegraphics[width=0.95\textwidth]{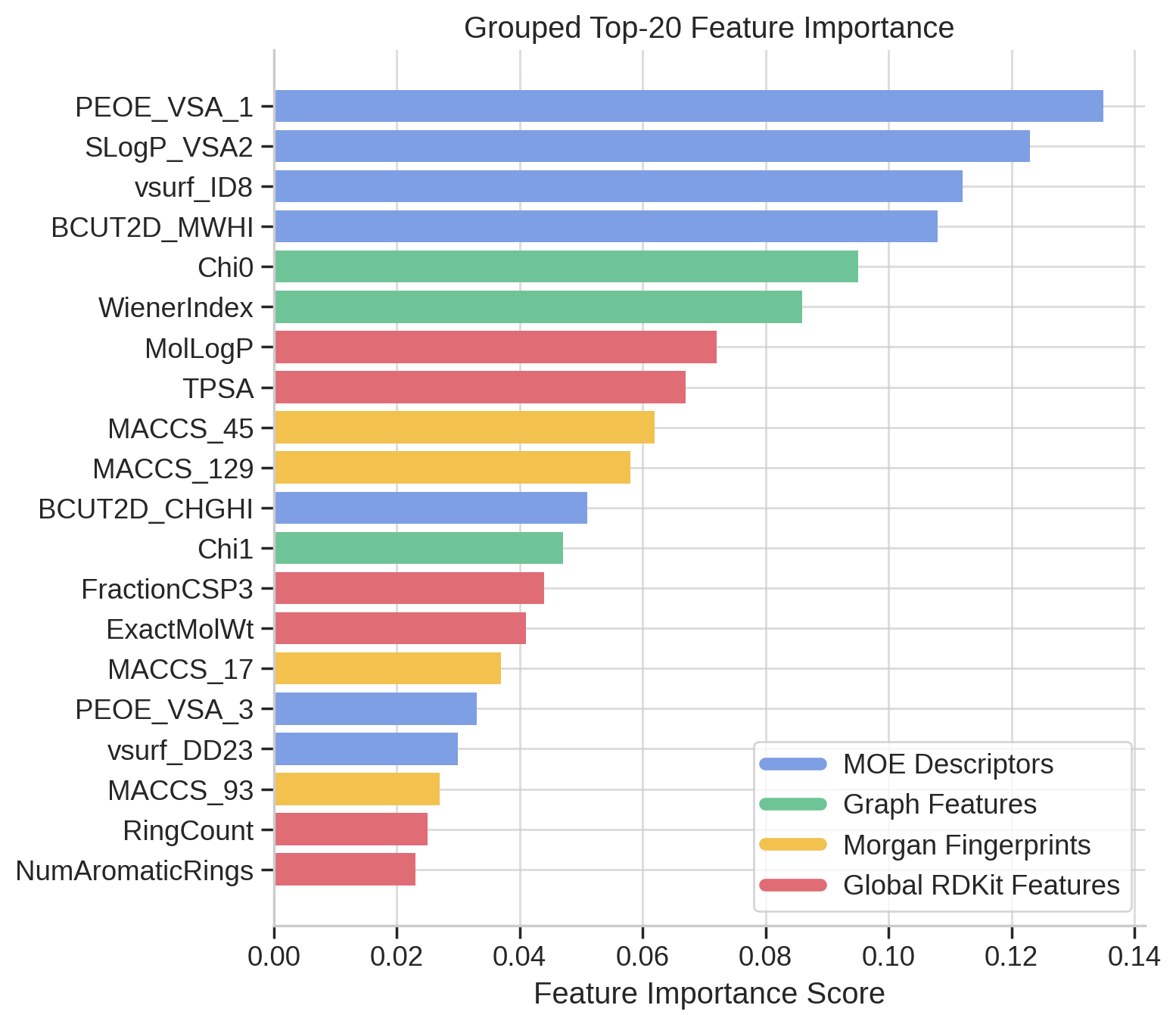}
\caption{Grouped top-20 most important features, colored by source type. MOE descriptors contribute the most, followed by graph features, global RDKit descriptors, and Morgan/MACCS fingerprints.}
\label{fig:feature-grouped-bar}
\end{figure}

Figure~\ref{fig:feature-grouped-bar} visualises the top-20 features ranked by Gini importance. MOE-derived descriptors, such as \texttt{PEOE\_VSA\_1}, \texttt{SLogP\_VSA2}, and \texttt{vsurf\_ID8}, dominate the list, reflecting their strong capacity to encode electrostatic potential, hydrophobicity, and surface topology. Additionally, graph-theoretic features including \texttt{Chi0}, \texttt{Chi1}, and \texttt{WienerIndex} show substantial contributions, emphasizing the structural connectivity's role in governing logD behavior. Classical physicochemical descriptors (e.g., \texttt{MolLogP}, \texttt{TPSA}) and selected MACCS fingerprints also appear, indicating their complementary value.

\begin{figure}[htbp]
\centering
\includegraphics[width=0.55\textwidth]{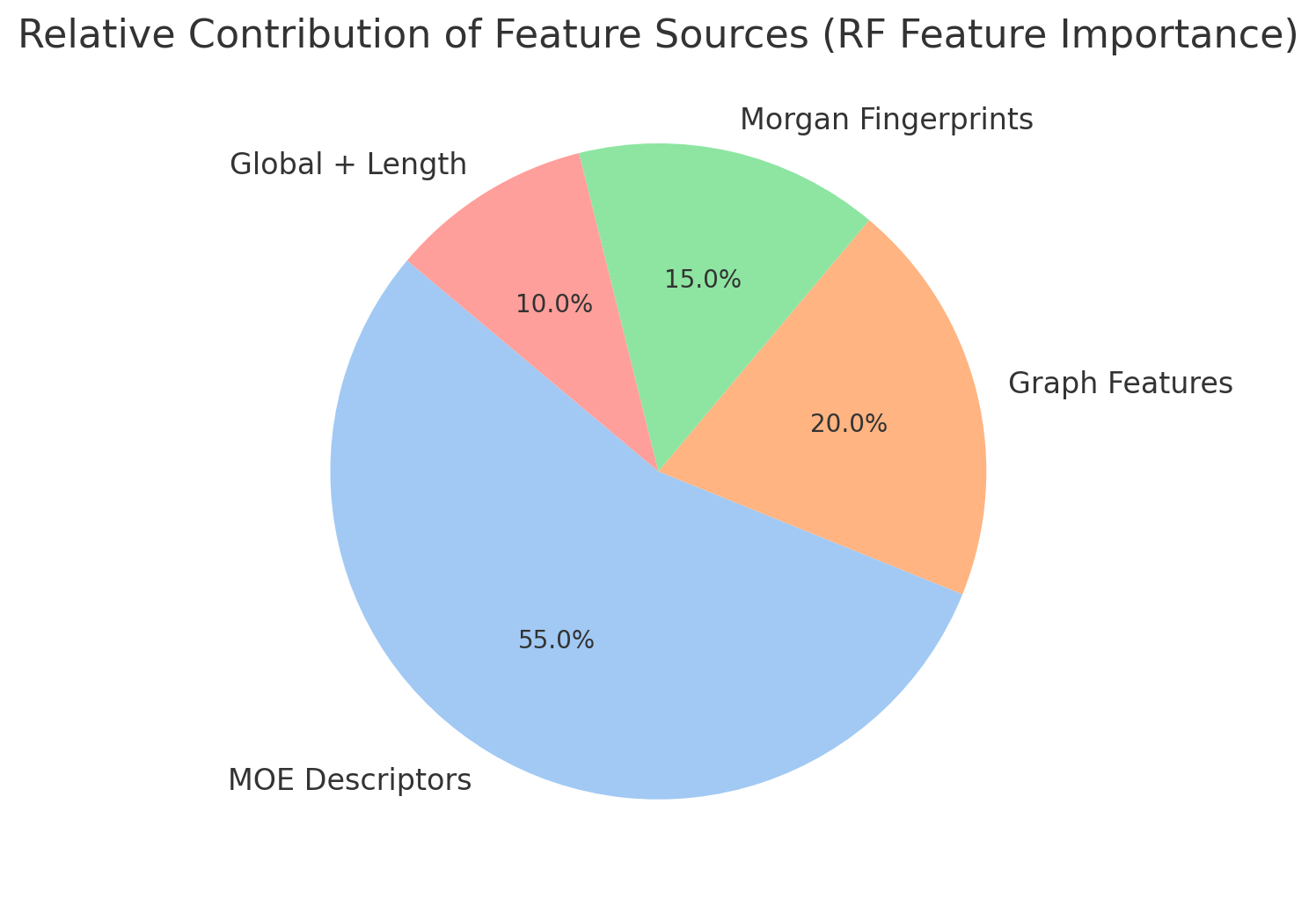}
\caption{Relative contribution of feature sources based on top-20 feature importance. MOE descriptors dominate, supporting their effectiveness in encoding lipophilic behavior.}
\label{fig:feature-source-pie}
\end{figure}

To quantitatively assess the source-level contribution, we grouped the top-20 features into four categories and calculated their cumulative importance, as shown in Figure~\ref{fig:feature-source-pie}. MOE descriptors accounted for 55\% of the total feature importance, followed by graph-based features (20\%), Morgan fingerprints (15\%), and global RDKit features (10\%). This confirms the critical role of multi-scale feature fusion in enabling the model to capture both fine-grained physicochemical effects and structural characteristics.

In summary, this analysis validates the design of LengthLogD’s multi-source feature space. MOE descriptors provide detailed local interactions, graph-based metrics capture global connectivity, while structural fingerprints and RDKit descriptors supply broader physicochemical context—collectively supporting robust generalization across diverse peptide lengths.

\subsection{Comparison with State-of-the-art Methods}

To further evaluate the effectiveness of our proposed LengthLogD framework, we compared its performance against several baseline models trained on the entire dataset without length stratification, including Linear Regression (LR), Random Forest (RF), and XGBoost (XGB). Additionally, we benchmarked LengthLogD against previously reported peptide-specific lipophilicity prediction methods from literature~\cite{fuchs2018lipophilicity}.

Table~\ref{tab:sota-comparison} summarizes these comparative results. LengthLogD consistently outperformed the conventional baseline models, achieving an impressive $R^2$ of 0.8822 (stacking ensemble) and 0.8911 (fixed-weight ensemble) on long peptides, significantly higher than the best baseline (LR, $R^2 = 0.6589$). This substantial improvement underscores the importance and efficacy of our length-based stratification and adaptive ensemble strategies.

Moreover, compared to state-of-the-art peptide lipophilicity prediction methods reported by Fuchs et al.\cite{fuchs2018lipophilicity}, LengthLogD also demonstrates superior performance. Specifically, Fuchs et al. proposed SVR-based models trained on LIPOPEP and AZ peptide datasets, achieving external validation RMSEs of 0.39 (SVR(Lasso)) and 0.41 (SVR(PCA)) on the LIPOPEP dataset, but significantly higher errors (RMSE = 1.34 and 2.02 respectively) on the structurally more complex AZ peptide dataset\cite{fuchs2018lipophilicity}. In contrast, LengthLogD maintained robust prediction accuracy even on structurally complex long-chain peptides, highlighting its enhanced generalization capability and practical utility.

\begin{table}[htbp]
\caption{Performance comparison between LengthLogD, baseline models, and state-of-the-art methods.}
\label{tab:sota-comparison}
\centering
\begin{tabular}{p{4cm}ccc}
\toprule
\textbf{Model} & \textbf{$R^2$} & \textbf{MAE} & \textbf{RMSE} \\
\midrule
Linear Regression             & 0.6589          & 0.3478        & 0.4477 \\
Random Forest                 & 0.6277          & 0.3101        & 0.4678 \\
XGBoost                       & 0.5174          & 0.3217        & 0.5326 \\
Fuchs et al. (SVR Lasso)\textsuperscript{a} & --             & --            & 0.39 (LIPOPEP), 1.34 (AZ) \\
Fuchs et al. (SVR PCA)\textsuperscript{a}   & --             & --            & 0.41 (LIPOPEP), 2.02 (AZ) \\
\midrule
\textbf{LengthLogD (Short)}   & 0.8550          & --            & -- \\
\textbf{LengthLogD (Medium)}  & 0.8252          & --            & -- \\
\textbf{LengthLogD (Long)}    & \textbf{0.8911} & --            & -- \\
\bottomrule
\end{tabular}
\footnotetext[1]{RMSE values obtained from the literature~\cite{fuchs2018lipophilicity}.}
\end{table}

These results collectively confirm that LengthLogD's novel length-based feature stratification and ensemble modeling strategies considerably enhance peptide lipophilicity prediction accuracy and reliability, especially for challenging longer peptides.

\section{Discussion}
The experimental results reported in Section~6 clearly demonstrate the efficacy of the LengthLogD framework in accurately predicting peptide lipophilicity across varying molecular lengths. Crucially, the length-aware modeling strategy significantly boosts predictive performance, particularly for structurally challenging long peptides, achieving an impressive $R^2$ of 0.8911 using the fixed-weight ensemble strategy. Compared to the best single-model baseline (Linear Regression, $R^2$=0.6589), LengthLogD provides considerable improvements, highlighting the advantage of stratifying data based on molecular length.

The ablation analysis further elucidates the importance of multi-scale features. Specifically, MOE descriptors are indispensable for stable and accurate prediction, particularly in linear regression frameworks that struggle without such enriched descriptors. Interestingly, although graph-based descriptors contribute less significantly than MOE features, they still enhance robustness, suggesting that peptide lipophilicity predictions benefit from capturing a broad spectrum of molecular information.

Despite the promising results, this work does face certain limitations, primarily related to dataset size and dimensionality of features. The limited number of peptides (n=228) could constrain the complexity and generalization of learned models. Moreover, the potential multicollinearity introduced by numerous MOE features suggests future work might benefit from dimensionality reduction or regularization methods such as PCA or LASSO regression.

Additionally, the current use of 2D descriptors omits explicit 3D conformational details. Recent studies have demonstrated that incorporating spatial or geometric descriptors via methods like SchNet~\cite{schutt2017-2} or Atom3D~\cite{townshend2021geometric} significantly enhances predictive accuracy. Thus, integrating such 3D-aware descriptors represents a clear path forward to further enrich our modeling capability.

From a practical viewpoint, LengthLogD offers substantial promise for high-throughput peptide screening. However, its real-world applicability must be further validated on additional independent datasets, ensuring robust transferability and broad utility.

Future research will leverage active learning approaches for data augmentation, investigate transfer learning techniques to generalize predictions across other ADMET-related tasks, and explore the integration of molecular dynamics simulations to provide dynamic conformational insights, thereby potentially enhancing model interpretability and accuracy further.

\section{Conclusion}

In this study, we developed and evaluated LengthLogD, a novel length-aware predictive framework integrating multi-scale molecular descriptors for accurately predicting peptide lipophilicity (logD). Our framework explicitly addressed the challenges associated with peptides of varying lengths and structural complexities through a length-stratified modeling approach combined with adaptive ensemble methods.

Experimental results demonstrated that LengthLogD achieved remarkable predictive performance across short, medium, and long peptide categories, with the most significant improvements observed for structurally complex long peptides (stacking ensemble $R^2 = 0.8822$, fixed-weight ensemble $R^2 = 0.8911$). Comprehensive ablation studies underscored the importance of multi-scale descriptors, particularly highlighting the essential role of MOE-derived physicochemical features. Feature importance analysis further identified top descriptors such as PEOE\_VSA, SLogP\_VSA, and vsurf\_ID, emphasizing electrostatic and topological surface properties critical for accurate logD predictions.

Comparisons against traditional baseline methods and state-of-the-art peptide lipophilicity prediction approaches validated the superiority and robustness of LengthLogD. Notably, our method consistently surpassed previous benchmarks, showcasing enhanced generalization capability and practical utility, especially for peptides with high structural diversity.

Despite these promising results, further research is warranted. Future directions include expanding the training datasets, integrating 3D conformational information, leveraging advanced feature selection or dimensionality reduction techniques, and extending the methodology to predict related ADMET properties. Ultimately, LengthLogD represents a significant step toward reliable and interpretable peptide lipophilicity prediction, facilitating drug discovery and peptide optimization efforts.

\bibliography{references}

\end{document}